# Development of an Advisory System for Parking of a Car and Trailer

Xincheng Cao, Haochong Chen, Bilin Aksun Guvenc, Levent Guvenc
Ohio State University

Shihong Fan, John Harber, Brian Link, Peter Richmond, Dokyung Yim
Hyundai America Technical Center Inc.

## Abstract

Trailer parking is a challenging task due to the unstable nature of the vehicle-trailer system in reverse motion and the unintuitive steering actions required at the vehicle to accomplish the parking maneuver. This paper presents a strategy to tackle this kind of maneuver with an advisory graphic aid to help the human driver with the task of manually backing up the vehicle-trailer system. A kinematic vehicle-trailer model is derived to describe the low-speed motion of the vehicle-trailer system, and its inverse kinematics is established by generating an equivalent virtual trailer axle steering command. The advisory system graphics is generated based on the inverse kinematics and displays the expected trailer orientation given the current vehicle steer angle and configuration (hitch angle). Simulation study and animation are set up to test the efficacy of the approach, where the user can select both vehicle speed and vehicle steering angle freely, which allows the user to stop the vehicle-trailer system and experiment with different steering inputs to see their effect on the predicted trailer motion before proceeding with the best one according to the advisory graphics, hence creating a series of piecewise continuous control actions similar to how manual trailer reverse parking is usually carried out. The advisory graphics proves to provide the driver with an intuitive understanding of the trailer motion at any given configuration (hitch angle).

## Introduction

Following many limited scale autonomous shuttle deployments [1], [2], public road deployments of autonomous driving have increased in recent years. The level of autonomy in a vehicle is classified using SAE's levels of autonomy [3]. There have also been series production vehicles with SAE level L2+ and L3 autonomy. The most basic function of an autonomous or automated driving vehicle is to plan and track its own path [4]. Higher levels of autonomy mean that more of the driving tasks like parking need to be automated [5]. The driving tasks that need to be automated in the future also include operation of a vehicle with a trailer and parking of a vehicle with a trailer.

Trailer reverse parking is a tricky maneuver to complete. Several reasons contribute to the difficulty in this maneuver. One is the inherently unstable nature of the vehicle-trailer system in reverse motion. Another reason is that depending on the system hitch angle (the angle between vehicle heading and trailer heading), different steering inputs at the vehicle steerable axle are required to orientate the trailer the same way. To help inexperienced drivers to handle this type of maneuver, an advisory system can be an easy-to-implement and cost-effective approach, motivating the analysis performed in this paper.

In order to construct an advisory system, a vehicle-trailer model is needed to describe system behaviors. In general, there are two types of models used for vehicle-trailer systems: the dynamics models that involve interactions of forces acting on the system components and the kinematic models that focus on geometric relationships.

The dynamic models can be derived using various methods. In [6], for example, a tractor-double-trailer system is analyzed. One of the trailers is a semi-trailer, while the other a full trailer, resulting in the whole system being treated as a four-component multi-body system. The Newton-Euler approach is used to derive the system model also using a linear tire model. Similarly, [7] provides the derivation of a dynamic semi-tractor-trailer model using the Newton-Euler method while exploring two different types of tire side slip angle models. Another method typically used in the derivation of dynamic models is the use of Euler-Lagrange equations. For instance, [8] provides the derivation procedure for the dynamic models of articulated commercial vehicles under a number of different configurations using the Euler-Lagrange method. Similarly, [9] derives the dynamics of an articulated truck using the same approach, treating the system as a five-axle articulated bicycle model. Additionally, [10] arrives at a nonlinear 4-DOF (degree-of-freedom) model using the Lagrange equations as well, and proceeds with model reduction to shrink the DOF from 4 to 3. Some comparative studies also exist to evaluate the accuracy of the dynamic models. For example, [11] compares the models described in [6], [7], [8], [9] under the five-axle configuration, while reference [12] compares four different dynamic vehicle-trailer models: linear 3-DOF, nonlinear 4-DOF, nonlinear 6-DOF and a high-fidelity nonlinear 21-DOF CarSim model.

Apart from the various versions of the dynamic models, one can also derive simpler kinematic models to analyze the behaviors of a vehicle-trailer system. Such models are most suited for low-speed ranges where parking maneuvers are typically carried out, since they are most accurate when excessive tire deformations are not present. Reference [13] describes the procedure to derive a kinematic vehicle-trailer model for a one-trailer setup partially using the concept of instantaneous center of rotation for the vehicle. Reference [14], on the other hand, uses kinematic constraint equations to derive the kinematic model for a system containing a tractor, a semi-trailer and a drawbar trailer. Meanwhile, reference [15] describes the kinematic model of a generic n-trailer setup while discussing the flatness property of the system.

For the motion control of the vehicle-trailer system, multiple approaches are utilized in existing research. One approach is to use two-stage controllers. For example, references [11] and [12] present a a high-level controller that generates the desired hitch angle and a low-level controller to follow that. Similarly, reference [18] describes a high-level controller that generates the desired curvature and a low-level controller to track such curvature. Additionally, reference [19] presents a control system that first calculates the expected tractor and trailer yaw rates before generating a steering input to follow them. Another approach treats the last trailer in the vehicle-trailer chain as the virtual tractor, and the required steer angle at the trailer for its proper orientation is translated through geometric relations to the steerable axle of the vehicle. References [14], [20] and [21] describe this method in detail. Other approaches include variations of model predictive control such as the distributed nonlinear MPC used in [22], model-reference control such as the nonlinear control system described in [23] and Lyapunov-based trajectory tracking controller such as the force-compensation controller proposed in [24].

Despite all its benefits, fully automated systems for vehicle-trailer parking are challenging to implement. In contrast, an advisory system, with the human driver in control, is an easier-to-implement and cost-effective alternative to handle the challenges of vehicle-trailer parking maneuvers. This paper hence aims to provide the details of developing such an advisory system to aid the drivers in understanding the behaviors of the vehicle-trailer system so that they can carry out the parking maneuver with ease. The outline of the rest of the paper is as follows. The kinematic vehicle-trailer model is developed next. This is followed by an inverse kinematic formulation. The trailer parking advisory system is, then, introduced and its effectiveness is demonstrated using simulations. The paper ends with conclusions.

## Kinematic Vehicle-Trailer Model

This section presents the derivation of a generic kinematic vehicle-trailer system for the one-trailer case. The system under consideration is displayed schematically in Figure 1. The parameters of this model are listed in Table 1. Note that value selection of $L_H$ determines the configuration of the vehicle (semi-tractor or passenger vehicle), and its value can be negative. If we treat vehicle rear axle as the reference point, one can write the position equations of key points in Equation 1. Applying time derivatives, velocity equations can be obtained as displayed in Equation 2.

$$\begin{cases} \begin{cases} X_R \\ Y_R \end{cases} \\ \begin{cases} X_F = X_R + L \cdot \cos(\psi_1) \\ Y_F = Y_R + L \cdot \sin(\psi_1) \end{cases} \\ \begin{cases} X_H = X_R - L_H \cdot \cos(\psi_1) \\ Y_H = Y_R - L_H \cdot \sin(\psi_1) \end{cases} \\ \begin{cases} X_T = X_H - L_T \cdot \cos(\psi_2) \\ Y_T = Y_H - L_T \cdot \sin(\psi_2) \end{cases} \end{cases} \tag{1}$$

$$\begin{cases} \overrightarrow{V_R} \begin{cases} \dot{X}_R \\ \dot{Y}_R \end{cases} \\ \overrightarrow{V_F} \begin{cases} \dot{X}_F = \dot{X}_R - L \cdot \dot{\psi}_1 \cdot \sin(\psi_1) \\ \dot{Y}_F = \dot{Y}_R + L \cdot \dot{\psi}_1 \cdot \cos(\psi_1) \end{cases} \\ \overrightarrow{V_H} \begin{cases} \dot{X}_H = \dot{X}_R + L_H \cdot \dot{\psi}_1 \cdot \sin(\psi_1) \\ \dot{Y}_H = \dot{Y}_R - L_H \cdot \dot{\psi}_1 \cdot \cos(\psi_1) \end{cases} \\ \overrightarrow{V_T} \begin{cases} \dot{X}_T = \dot{X}_H + L_T \cdot \dot{\psi}_2 \cdot \sin(\psi_2) \\ \dot{Y}_T = \dot{Y}_H - L_T \cdot \dot{\psi}_2 \cdot \cos(\psi_2) \end{cases} \end{cases} \tag{2}$$

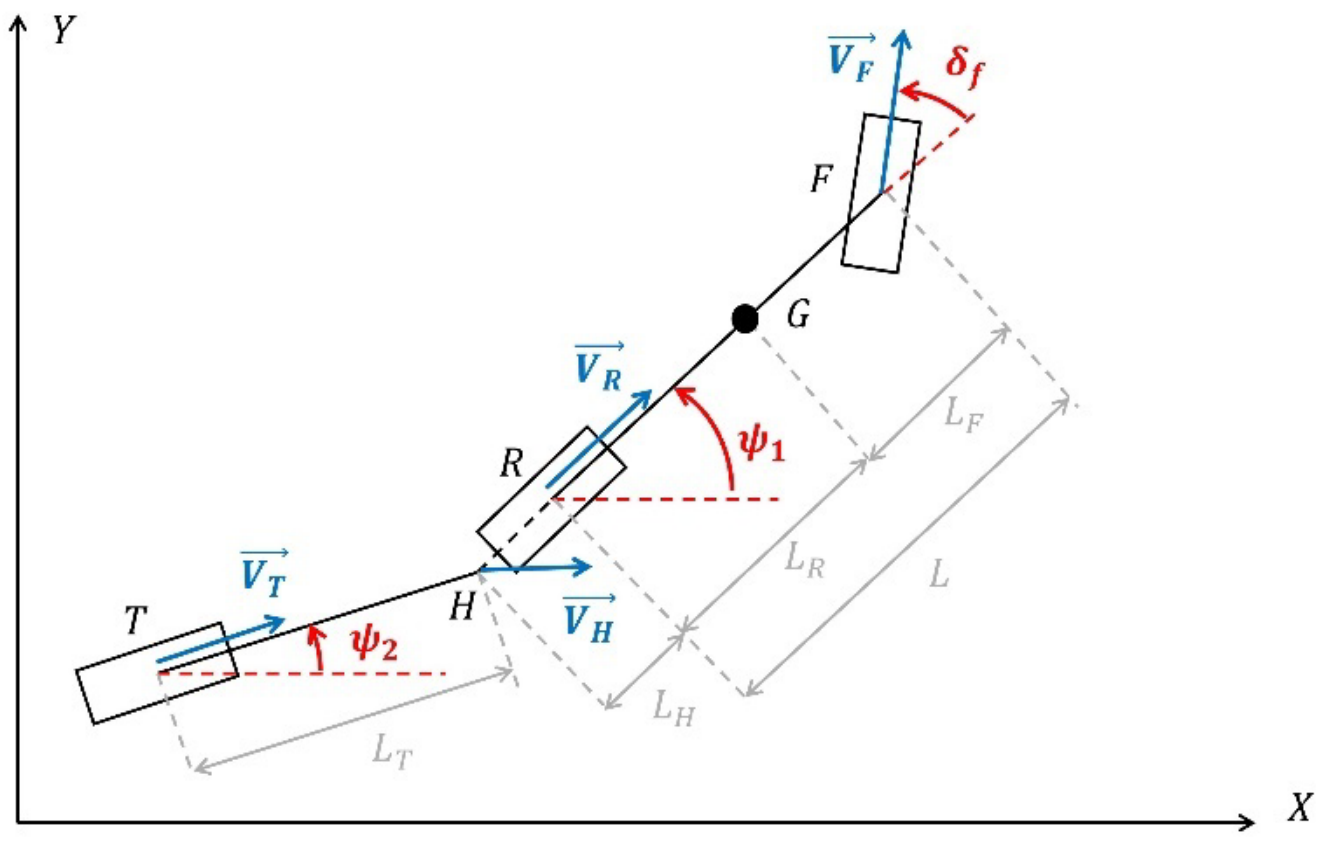

Figure 1. Kinematic vehicle-trailer model

Table 1. Parameters of kinematic tractor-trailer model

| Model Parameter | Explanation |
|---|---|
| $L$ | Wheelbase of the tractor vehicle |
| $L_F$ | Distance between vehicle CG and front axle |
| $L_R$ | Distance between vehicle CG and rear axle |
| $L_H$ | Distance between vehicle rear axle and trailer hitch |
| $L_T$ | Distance between trailer axle and trailer hitch |
| $\delta_f$ | Vehicle front wheel steer angle |
| $\psi_1$ | Vehicle yaw angle |
| $\psi_2$ | Trailer yaw angle |
| $\overrightarrow{V_F}$ | Vehicle front axle velocity |
| $\overrightarrow{V_R}$ | Vehicle rear axle velocity |
| $\overrightarrow{V_H}$ | Trailer hitch velocity |
| $\overrightarrow{V_T}$ | Trailer axle velocity |

In Figure 1, note that the velocities of the vehicle and trailer wheels are aligned with their respective orientation. This is due to the assumption that at low speed where parking maneuver occurs, there is no tire side slip so that the direction of tire travel follows the direction in which the tire is pointing. This assumption will induce kinematic constraints in the model, and they can be represented in Equation 3. Note that the term $V_R$ is the magnitude of the vehicle rear axle velocity $\overrightarrow{V_R}$ and can either be positive or negative.

$$\begin{cases} \tan(\psi_1) = \frac{V_R \cdot \sin(\psi_1)}{V_R \cdot \cos(\psi_1)} = \frac{\dot{Y}_R}{\dot{X}_R} \\ \tan(\psi_1 + \delta_f) = \frac{\dot{Y}_F}{\dot{X}_F} \quad \text{, where} \begin{cases} V_R \\ \delta_f \end{cases} \text{are model inputs} \\ \tan(\psi_2) = \frac{\dot{Y}_T}{\dot{X}_T} \end{cases} \tag{3}$$

Combining the velocity equations with the kinematic constraint equations and simplifying, one can obtain Equations 4.

$$\begin{cases} \tan(\psi_1 + \delta_f) = \frac{V_R \cdot \sin(\psi_1) + L \cdot \dot{\psi}_1 \cdot \cos(\psi_1)}{V_R \cdot \cos(\psi_1) - L \cdot \dot{\psi}_1 \cdot \sin(\psi_1)} \\ \tan(\psi_2) = \frac{\dot{Y}_T}{\dot{X}_T} = \frac{V_R \cdot \sin(\psi_1) - L_H \cdot \dot{\psi}_1 \cdot \cos(\psi_1) - L_T \cdot \dot{\psi}_2 \cdot \cos(\psi_2)}{V_R \cdot \cos(\psi_1) + L_H \cdot \dot{\psi}_1 \cdot \sin(\psi_1) + L_T \cdot \dot{\psi}_2 \cdot \sin(\psi_2)} \end{cases} \tag{4}$$

By simplifying individual components in Equation 4, the expressions of the kinematic tractor-trailer model can be obtained as Equation 5. Note that the term $V_T$ denotes the magnitude of trailer axle velocity $\overrightarrow{V_T}$ and is a function of the two inputs as well as the hitch angle which can be obtained by subtracting the previous two states. This results in

$$\begin{cases} [I]: \dot{\psi}_1 = \frac{V_R}{L}\tan(\delta_f) \\ [II]: \dot{\psi}_2 = \frac{V_R}{L_T}[\sin(\Delta\psi) - \frac{L_H}{L}\cos(\Delta\psi)\tan(\delta_f)], \\ [III]: V_T = V_R[\cos(\Delta\psi) + \frac{L_H}{L}\sin(\Delta\psi)\tan(\delta_f)] \end{cases} \quad (5)$$

where $\Delta\psi = \psi_1 - \psi_2$ is the hitch angle. To observe the behaviors of this kinematic vehicle-trailer model, a simulation routine is established based on Equation 5, the flowchart of which is displayed in Figure 2. Based on this simulation, animation can be constructed to show the behaviors of the system more intuitively by applying frame transformations with simulation outputs. Figure 3 provides a snippet of this animation.

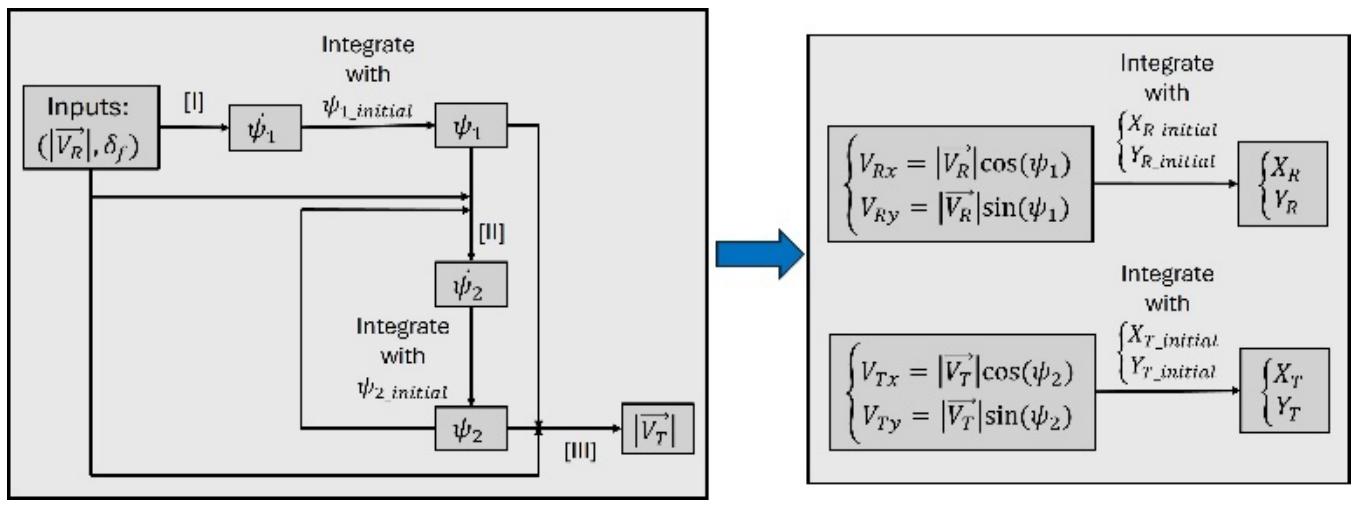

Figure 2. Kinematic tractor-trailer model simulation flowchart

## Inverse Kinematics

In reverse motion, the vehicle-trailer system demonstrates some unintuitive yaw behaviors for inexperienced drivers. As a result, it is helpful to first view the trailer as a standalone vehicle and work out the 'virtual' steering angle at the trailer that would be required to orientate the trailer properly, before mapping this 'virtual' angle to vehicle steer angle at the vehicle steerable axle through kinematic derivation. This section demonstrates the procedure to arrive at this relationship between the 'virtual' steer angle at the trailer and the actual steer angle at the vehicle.

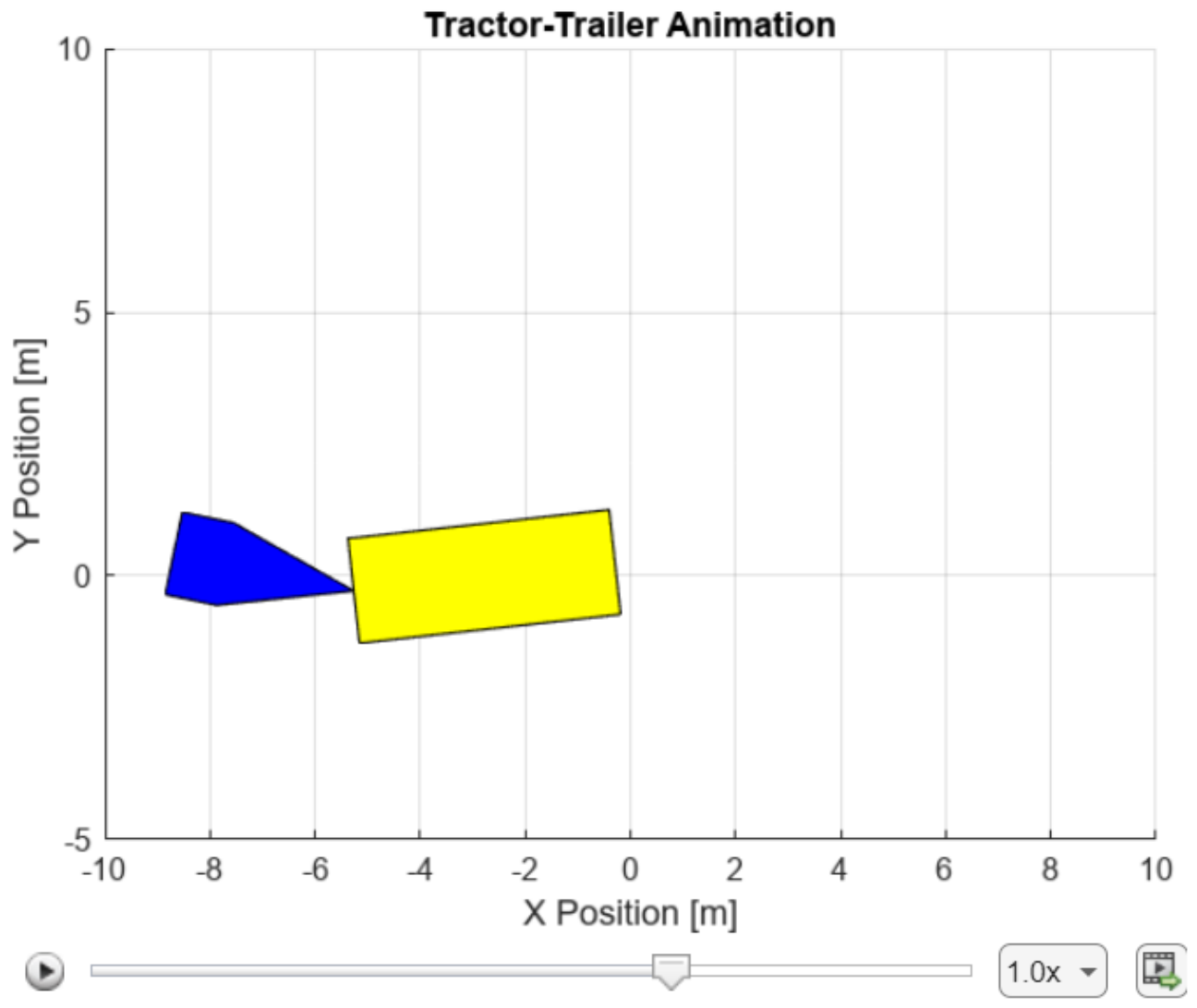

Figure 3. Snippet of tactor-trailer system animation

It should first be remarked that the inverse kinematic vehicle-trailer model is identical to the model illustrated in Figure 1 except for the addition of the 'virtual' steerable axle at the trailer hitch, as illustrated in Figure 4 that shows only the trailer part of the model to reduce visual clutter. With this virtual steerable axle, the trailer can be regarded as a standalone vehicle and its steer angle is denoted as $\delta_T$.

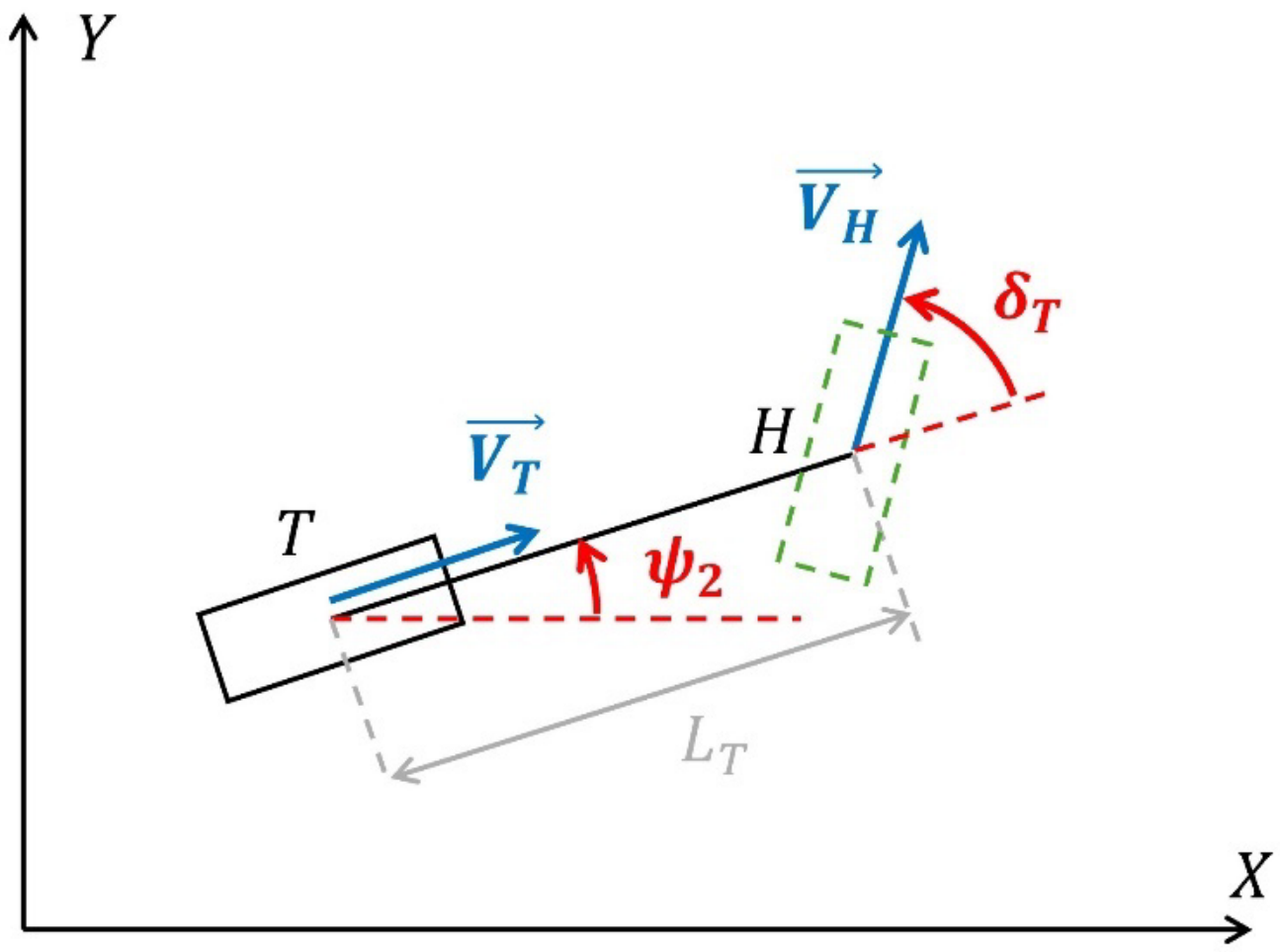

Figure 4. Inverse kinematic trailer-tractor model: trailer part

The process to derive the inverse kinematic model is also similar to the process outlined in the above section, except we now regard the trailer axle as the refence point. Hence, one can obtain the position equations in Equation 6 and velocity equations in Equation 7 by applying time derivative to Equation 6.

$$\begin{cases} \begin{cases} X_T \\ Y_T \end{cases} \\ \begin{cases} X_H = X_T + L_T \cdot \cos(\psi_2) \\ Y_H = Y_T + L_T \cdot \sin(\psi_2) \end{cases} \\ \begin{cases} X_R = X_H + L_H \cdot \cos(\psi_1) \\ Y_R = Y_H + L_H \cdot \sin(\psi_1) \end{cases} \\ \begin{cases} X_F = X_R + L \cdot \cos(\psi_1) \\ Y_F = Y_R - L \cdot \sin(\psi_1) \end{cases} \end{cases} \quad (6)$$

$$\begin{cases} \overrightarrow{V_T} \begin{cases} \dot{X}_T \\ \dot{Y}_T \end{cases} \\ \overrightarrow{V_H} \begin{cases} \dot{X}_H = \dot{X}_T - L_T \cdot \dot{\psi}_2 \cdot \sin(\psi_2) \\ \dot{Y}_H = \dot{Y}_T + L_T \cdot \dot{\psi}_2 \cdot \cos(\psi_2) \end{cases} \\ \overrightarrow{V_R} \begin{cases} \dot{X}_R = \dot{X}_H - L_H \cdot \dot{\psi}_1 \cdot \sin(\psi_1) \\ \dot{Y}_R = \dot{Y}_H + L_H \cdot \dot{\psi}_1 \cdot \cos(\psi_1) \end{cases} \\ \overrightarrow{V_F} \begin{cases} \dot{X}_F = \dot{X}_R - L \cdot \dot{\psi}_1 \cdot \sin(\psi_1) \\ \dot{Y}_F = \dot{Y}_R + L \cdot \dot{\psi}_1 \cdot \cos(\psi_1) \end{cases} \end{cases} \quad (7)$$

Applying kinematic constraints, the constraint equations can be obtained in Equation 8. Note that kinematic constraint at trailer virtual steer axle also applies, so the hitch velocity $\overrightarrow{V_H}$ is aligned with the 'virtual' steer angle $\delta_T$.

$$
\begin{cases}
\tan(\psi_2) = \frac{V_T \cdot \sin(\psi_2)}{V_T \cdot \cos(\psi_2)} = \frac{\dot{Y_T}}{\dot{X_T}} \\
\tan(\psi_2 + \delta_T) = \frac{\dot{Y_H}}{\dot{X_H}} \\
\tan(\psi_1) = \frac{\dot{Y_R}}{\dot{X_R}} \\
\tan(\psi_1 + \delta_f) = \frac{\dot{Y_F}}{\dot{X_F}}
\end{cases} \tag{8}
$$

Combining velocity equations and kinematic constraint equations, one can obtain Equation 9.

$$
\begin{cases}
\tan(\psi_2 + \delta_T) = \frac{V_T \cdot \sin(\psi_2) + L_T \cdot \dot{\psi_2} \cdot \cos(\psi_2)}{V_T \cdot \cos(\psi_2) - L_T \cdot \dot{\psi_2} \cdot \sin(\psi_2)} \\
\tan(\psi_1) = \frac{\dot{Y_R}}{\dot{X_R}} = \frac{V_T \cdot \sin(\psi_2) + L_T \cdot \dot{\psi_2} \cdot \cos(\psi_2) + L_H \cdot \dot{\psi_1} \cdot \cos(\psi_1)}{V_T \cdot \cos(\psi_2) - L_T \cdot \dot{\psi_2} \cdot \sin(\psi_2) - L_H \cdot \dot{\psi_1} \cdot \sin(\psi_1)} \\
\tan(\psi_1 + \delta_f) = \frac{V_T \cdot \sin(\psi_2) + L_T \cdot \dot{\psi_2} \cdot \cos(\psi_2) + L_H \cdot \dot{\psi_1} \cdot \cos(\psi_1) + L \cdot \dot{\psi_1} \cdot \cos(\psi_1)}{V_T \cdot \cos(\psi_2) - L_T \cdot \dot{\psi_2} \cdot \sin(\psi_2) - L_H \cdot \dot{\psi_1} \cdot \sin(\psi_1) - L \cdot \dot{\psi_1} \cdot \sin(\psi_1)}
\end{cases} \tag{9}
$$

Simplifying individual components in Equation 9, one can obtain Equation 10. It must be remarked that elements [I] and [II] of Equation 10 are the 'desired' yaw rates of the trailer and the vehicle, respectively, given a virtual steer angle $\delta_T$. Element [IV] of Equation 10 provides the mapping between actual steer angle at the vehicle front axle (the steerable axle) and the virtual steer angle at the trailer hitch. It should also be noted that, alternatively, one can equate element [I] of Equation 5 (actual vehicle yaw rate given control inputs $V_R$ and $\delta_f$) and element [II] of Equation 10 (desired vehicle yaw rate given $\delta_T$) and then plugging in element [III] of Equation 10 to obtain the actual-virtual steer angle mapping described in element [IV] of Equation 10 where $\Delta\psi = \psi_1 - \psi_2$.

$$
\begin{cases}
[I]: \dot{\psi_2} = \frac{V_T}{L_T} \tan(\delta_T) \\
[II]: \dot{\psi_1} = \frac{V_T}{L_H} [\sin(\Delta\psi) - \cos(\Delta\psi) \tan(\delta_T)] \\
[III]: V_T = \frac{V_R}{\cos(\Delta\psi) + \sin(\Delta\psi)\tan(\delta_T)} \\
[IV]: \delta_f = \operatorname{atan}\left(\frac{L}{L_H} \cdot \frac{\sin(\Delta\psi) - \cos(\Delta\psi)\tan(\delta_T)}{\cos(\Delta\psi) + \sin(\Delta\psi)\tan(\delta_T)}\right)
\end{cases} \tag{10}
$$

It should be noted that sign convention applies to the actual-virtual steer angle mapping equation ([IV] of Equation 10) depending on whether the vehicle-trailer system is in forward or backward motion. It should also be remarked that sign convention will be inverted if tractor configuration (position of trailer hitch relative to position of vehicle rear axle) changes. Equation 11 outlines the sign convention for each case, with [I] denoting trailer hitch behind vehicle rear axle, while [II] denotes trailer hitch in front of vehicle rear axle. This mapping can also be represented in reverse as illustrated in Equation 12. Note that sign convention applies to Equation 12 as well.

$$
\begin{cases}
[I] \begin{cases}
\delta_f = \operatorname{atan}\left(\frac{L}{L_H} \cdot \frac{\sin(\Delta\psi) - \cos(\Delta\psi)\tan(\delta_T)}{\cos(\Delta\psi) + \sin(\Delta\psi)\tan(\delta_T)}\right) \; in\ reverse \\
\delta_f = -\operatorname{atan}\left(\frac{L}{L_H} \cdot \frac{\sin(\Delta\psi) - \cos(\Delta\psi)\tan(\delta_T)}{\cos(\Delta\psi) + \sin(\Delta\psi)\tan(\delta_T)}\right) \; in\ forward
\end{cases} \\
[II] \begin{cases}
\delta_f = -\operatorname{atan}\left(\frac{L}{L_H} \cdot \frac{\sin(\Delta\psi) - \cos(\Delta\psi)\tan(\delta_T)}{\cos(\Delta\psi) + \sin(\Delta\psi)\tan(\delta_T)}\right) \; in\ reverse \\
\delta_f = \operatorname{atan}\left(\frac{L}{L_H} \cdot \frac{\sin(\Delta\psi) - \cos(\Delta\psi)\tan(\delta_T)}{\cos(\Delta\psi) + \sin(\Delta\psi)\tan(\delta_T)}\right) \; in\ forward
\end{cases}
\end{cases} \tag{11}
$$

$$
\delta_T = \operatorname{atan}\left(\frac{L \cdot \sin(\Delta\psi) - L_H \cdot \cos(\Delta\psi)\tan(\delta_f)}{L \cdot \cos(\Delta\psi) + L_H \cdot \sin(\Delta\psi)\tan(\delta_f)}\right) \tag{12}
$$

To demonstrate the effects of actual-virtual steer angle mapping, a simulation study is performed. The simulation routine described in the above section is expanded to include the inverse kinematics and is illustrated in Figure 5. A 'desired' $\delta_T$ profile is fed into the inverse kinematics calculation that invokes Equation 11, and the resulting vehicle steer angle required is fed into the vehicle-trailer model. An additional block is also included to calculate the 'actual' $\delta_T$ based on the outputs of the vehicle-trailer model, which is then compared to the 'desired' $\delta_T$ profile. The calculation procedure of the 'actual' $\delta_T$ is detailed in Equation 13.

$$
Actual\ \delta_T = \operatorname{atan}\left(\frac{\dot{Y_H}}{\dot{X_H}}\right) - \psi_2 = \operatorname{atan}\left(\frac{|\vec{V_T}|\sin(\psi_2) + L_T\dot{\psi_2}\cos(\psi_2)}{|\vec{V_T}|\cos(\psi_2) - L_T\dot{\psi_2}\sin(\psi_2)}\right) - \psi_2 \tag{13}
$$

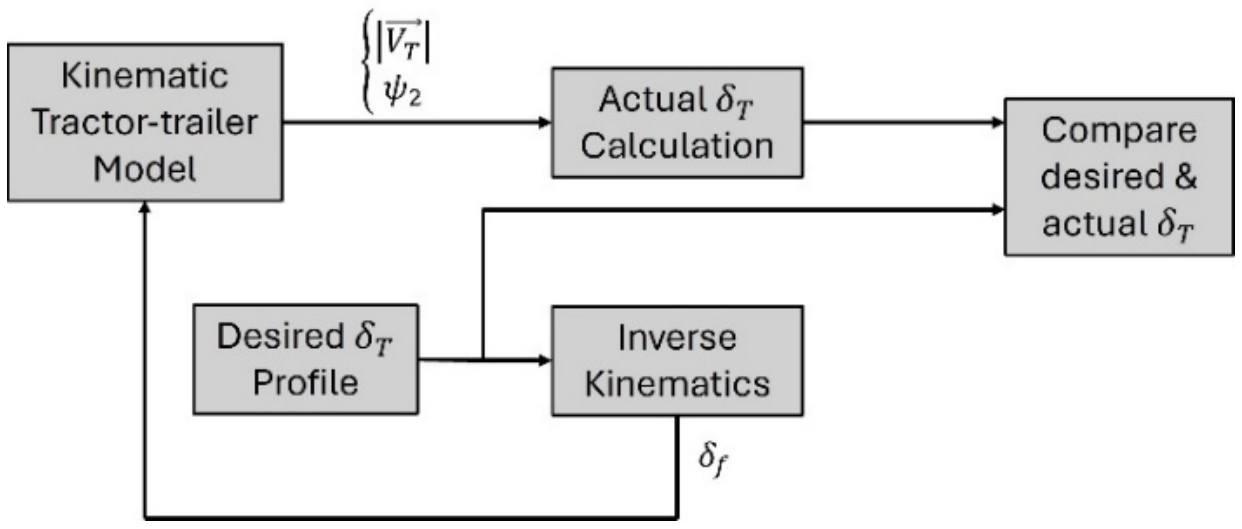

Figure 5. Inverse kinematics simulation study model structure

The simulation study is performed in both forward and backward motion with two vehicle configurations: trailer hitch behind the vehicle rear axle (i.e. passenger vehicle) and trailer hitch in front of the vehicle rear axle (i.e. semi-tractor). The results are illustrated in Figure 6 and Figure 7. It can be observed that while both vehicle configurations display the match of desired and actual $\delta_T$ in general trend, for each vehicle configuration, exact match is not achieved for either the forward or backward motion. This is due to the kinematic limitations of the vehicle configuration. Taking the case of forward motion with trailer hitch behind the vehicle rear axle as an example, vehicle yaw motion induces a local rotation centered around the rear axle that swings the tail (where the trailer hitch is located) in the direction opposite to the overall trailer hitch trajectory, causing the slight mismatch between actual and desired $\delta_T$. The extent of the mismatch depends on the distance between the trailer hitch and the vehicle rear axle, with zero distance eliminating the mismatch completely in both forward and backward motions.

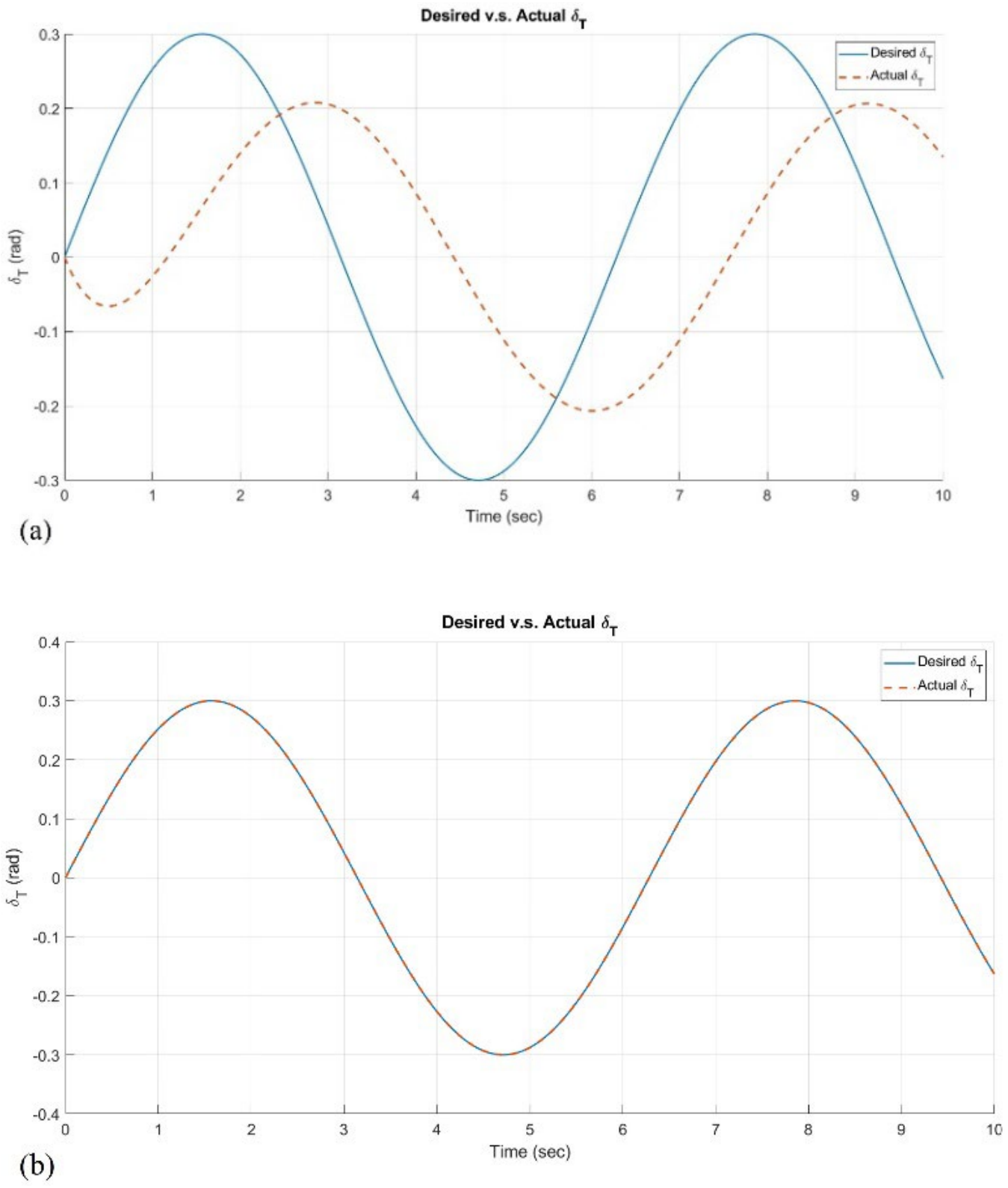

Figure 6. Desired virtual steer angle following performance for trailer hitch behind vehicle rear axle: (a) forward motion; (b) backward motion

## Trailer Parking Advisory System

If the trailer parking is to be done manually by human drivers, a good advisory system can significantly enhance the possibility of success, especially for inexperienced drivers. Existing parking advisory systems focus on providing aiding graphics to the driver so that the driver understands the current state of the vehicle (and trailer). Figure 8(a) shows an example of such an advisory graphics for the vehicle. In this example, there are two curves following the two tracks of the vehicle, which serve as the expected vehicle track trajectories given a certain vehicle speed and current steering input. It should be remarked that these two curves display the expected trajectories even when the vehicle is stationary. Figure 8(b) shows an example of existing advisory graphics for vehicle-trailer system. In this example, the graphics displays the current vehicle-trailer configuration (hitch angle) and the expected trailer orientation at a certain speed given the current steering input. This section aims to provide the construction details of an advisory system that achieve similar functionalities.

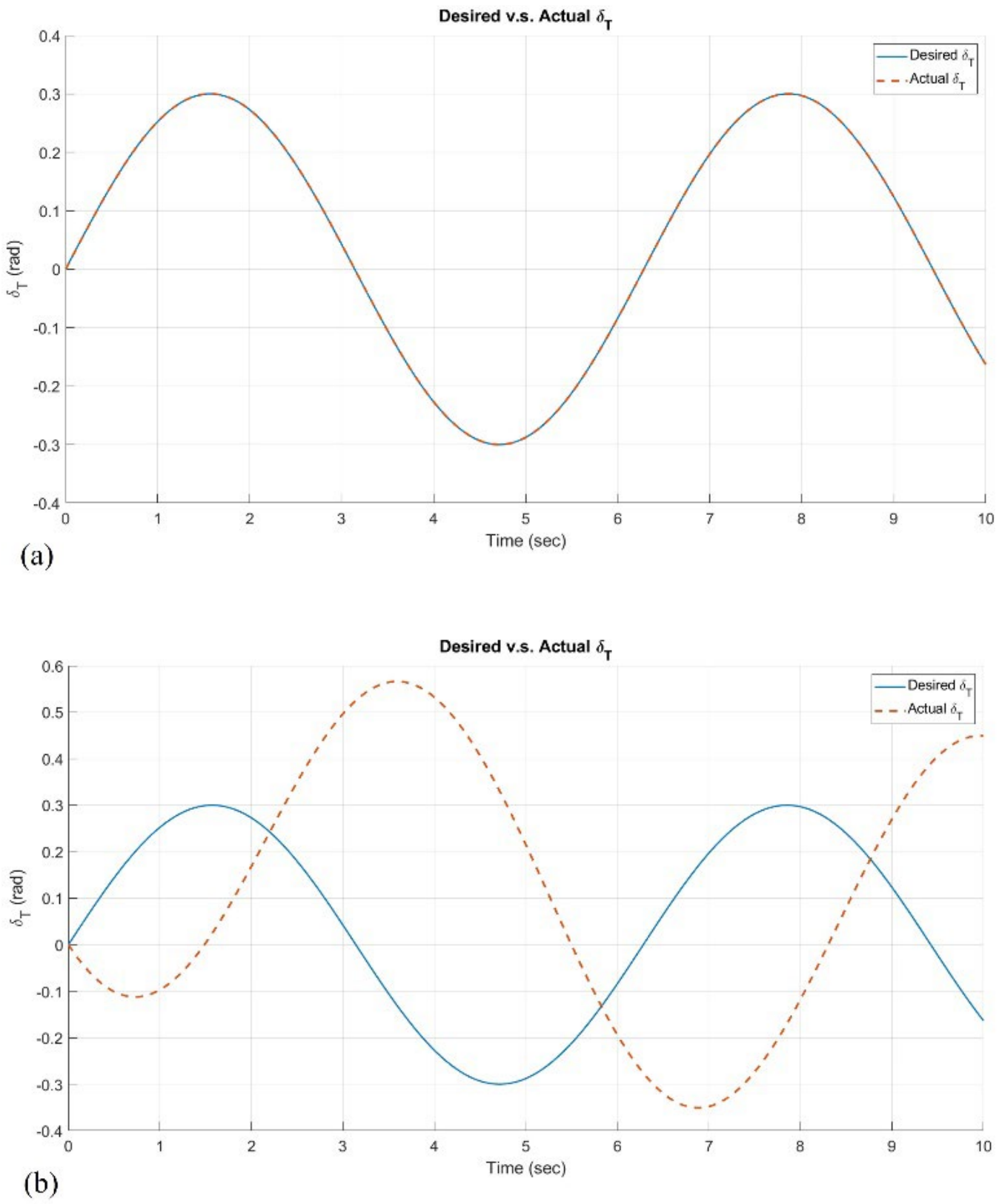

Figure 7. Desired virtual steer angle following performance for trailer hitch in front of vehicle rear axle: (a) forward motion; (b) backward motion

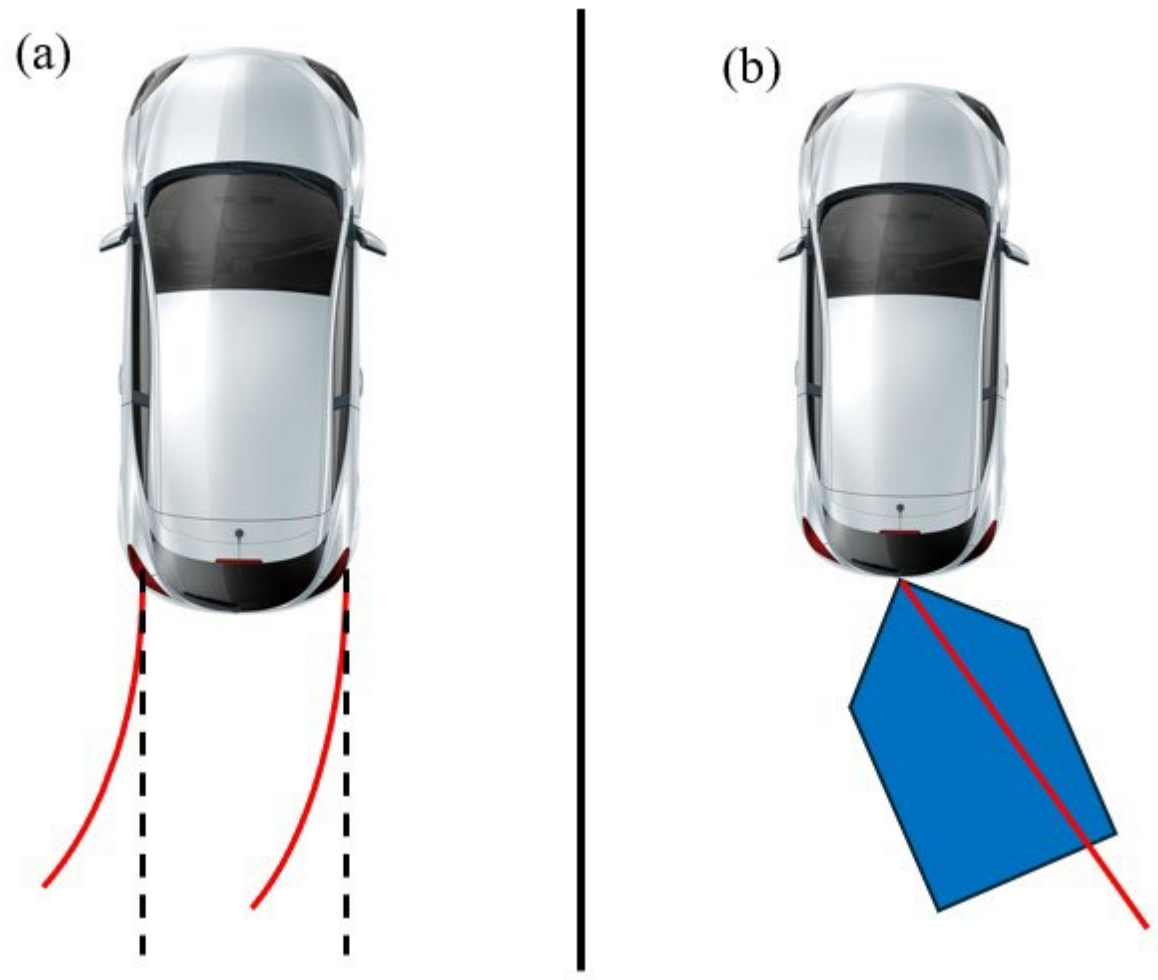

Figure 8. Existing advisory graphics: (a) for vehicle only; (b) for tractor-trailer system

The advisory system developed in this case is based on the simulation routine mentioned in the previous sections, where the kinematic model is used. The advisory graphics developed features a base layout consisting of two panels (left and right). The left panel is adapted from the animation mentioned in the above section and shows the top view of the vehicle-trailer system in the global coordinate frame. The vehicle-trailer system can move in this frame and additional components such as the outline of the parking space can also be added. The right panel shows the current configuration (hitch angle) of the vehicle-trailer system where the vehicle graphics remain stationary and the relative rotation of the trailer around the trailer hitch is displayed.

An additional straight line is added in the left panel of the graphics to denote the expected trailer orientation if the trailer is reversed for one second at a speed of 1 m/sec using current steering angle and configuration (hitch angle). To show this, expected trailer yaw angle $\psi_2$ should be calculated and Equation 12 as well as element [I] of Equation 10 can be used for this purpose. Figure 9 illustrates this process. For simulation, both vehicle speed and vehicle steering are introduced as user-generated inputs. The simulation pauses every second to request a new set of user inputs to update the vehicle speed and steering. Under this setup, the user has the option to stop the vehicle and try different steering angles to observe the expected future behavior of the trailer. Once the user is happy with the expected trailer motion with a certain steer input, they can use this steer angle and apply reverse motion by inputting negative speed. This will result in both the steering inputs and the vehicle speed to be piecewise continuous, which is more realistic, as this procedure is similar to how many human drivers perform reverse maneuvers, where they stop the vehicle to evaluate the situation and select a different steering angle before proceeding further. A snippet of this advisory system simulation is provided in Figure 10, and the plotted results are displayed for two different trailer wheelbases in Figure 11 ($L_T = 2.5m$) and Figure 12 ($L_T = 1.5m$), respectively. It is worth remarking that since the vehicle speed is user-defined, the user can also pull forward to straighten out the vehicle-trailer system and obtain a more favorable configuration (hitch angle) before backing up again. These results and their animations corresponding to real life trailer-parking assistance demonstrate the effectiveness of this proposed advisory system.

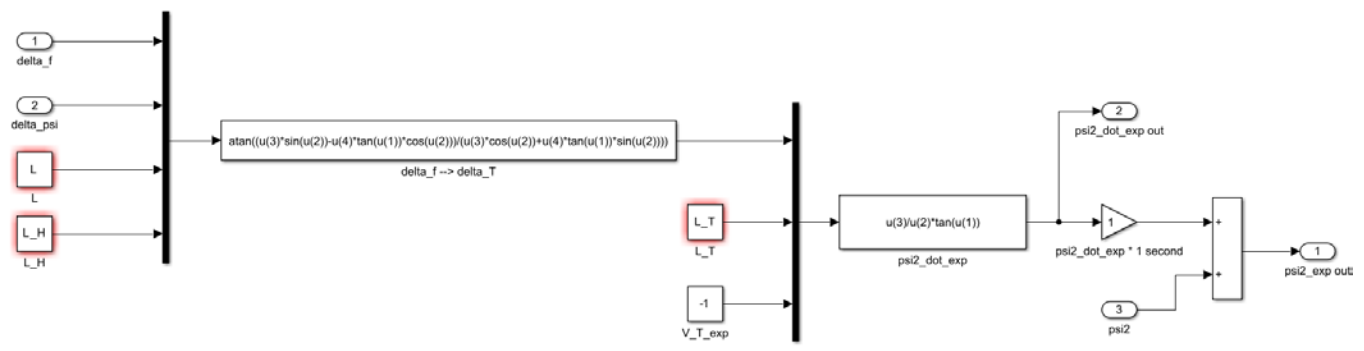

Figure 9. Advisory system calculation

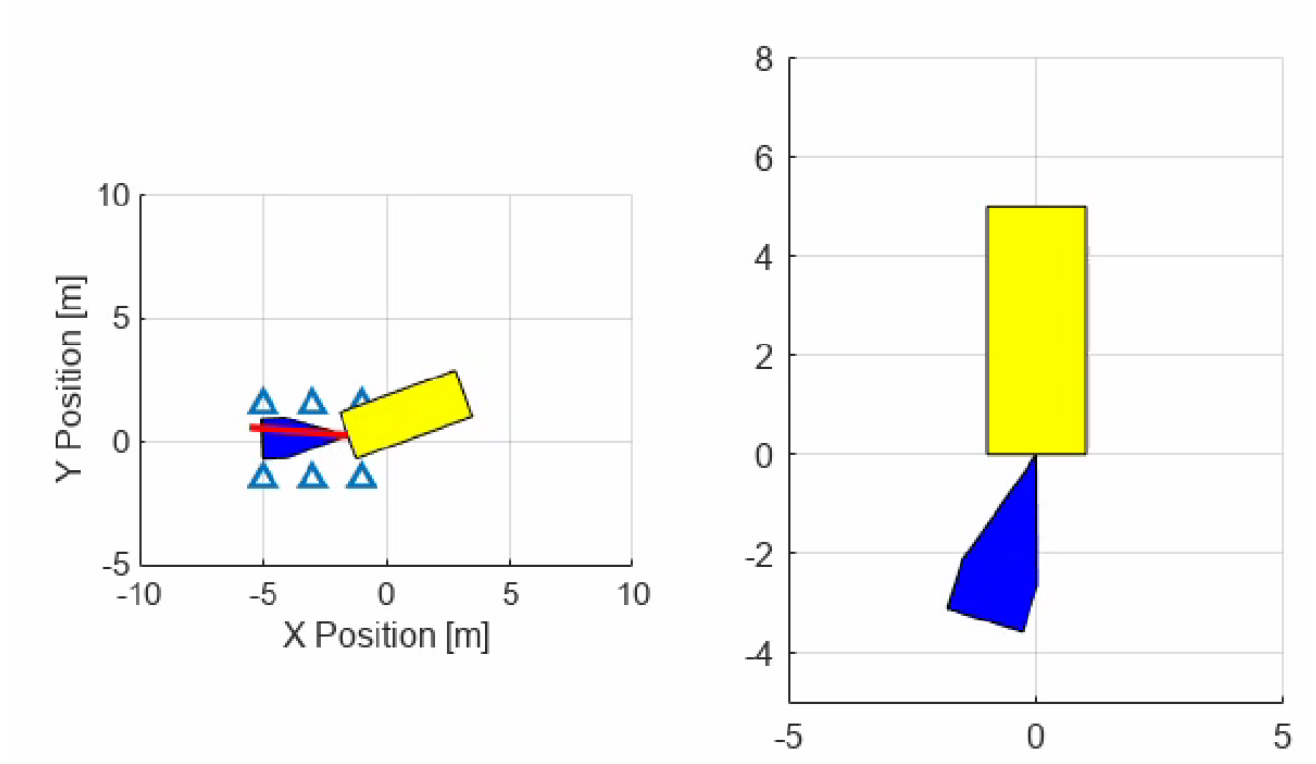

Figure 10. Advisory system graphics

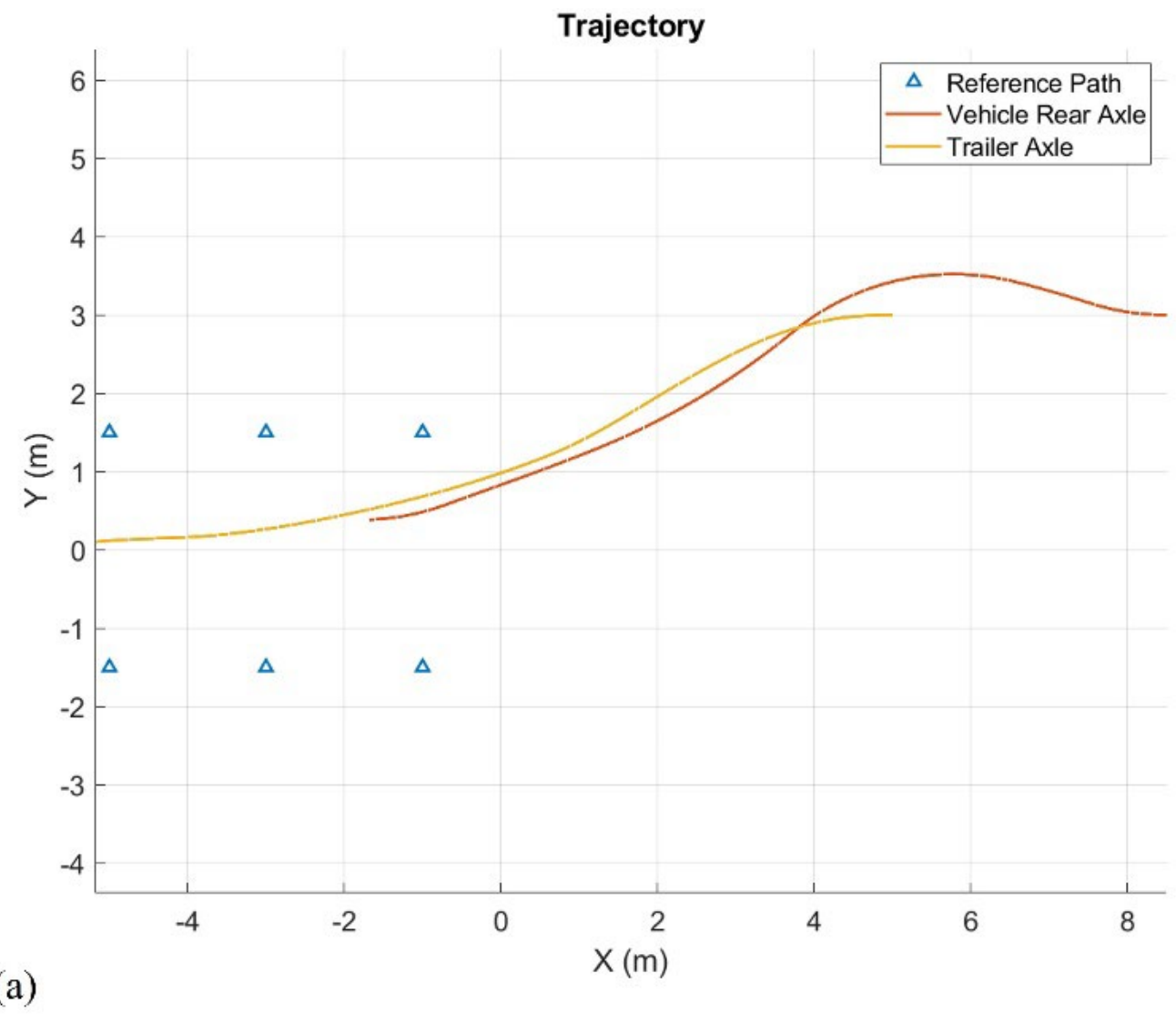

(a)

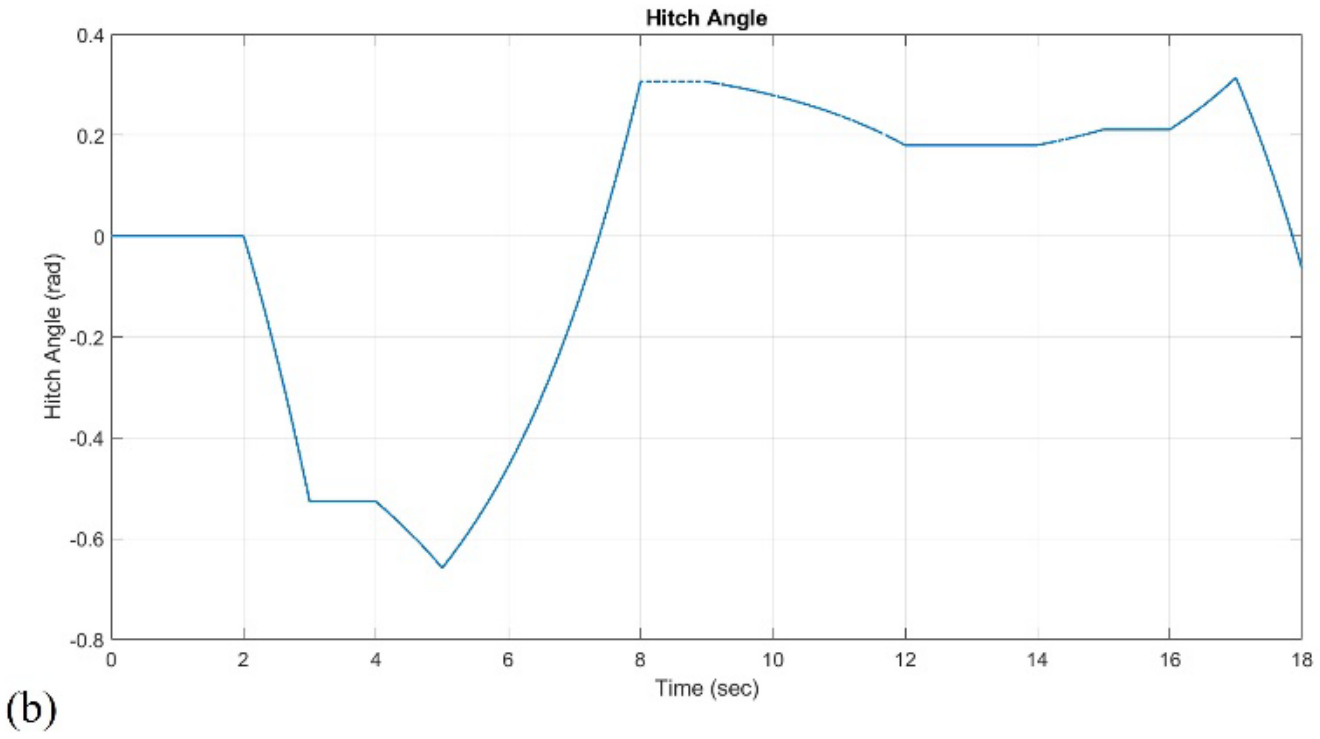

(b)

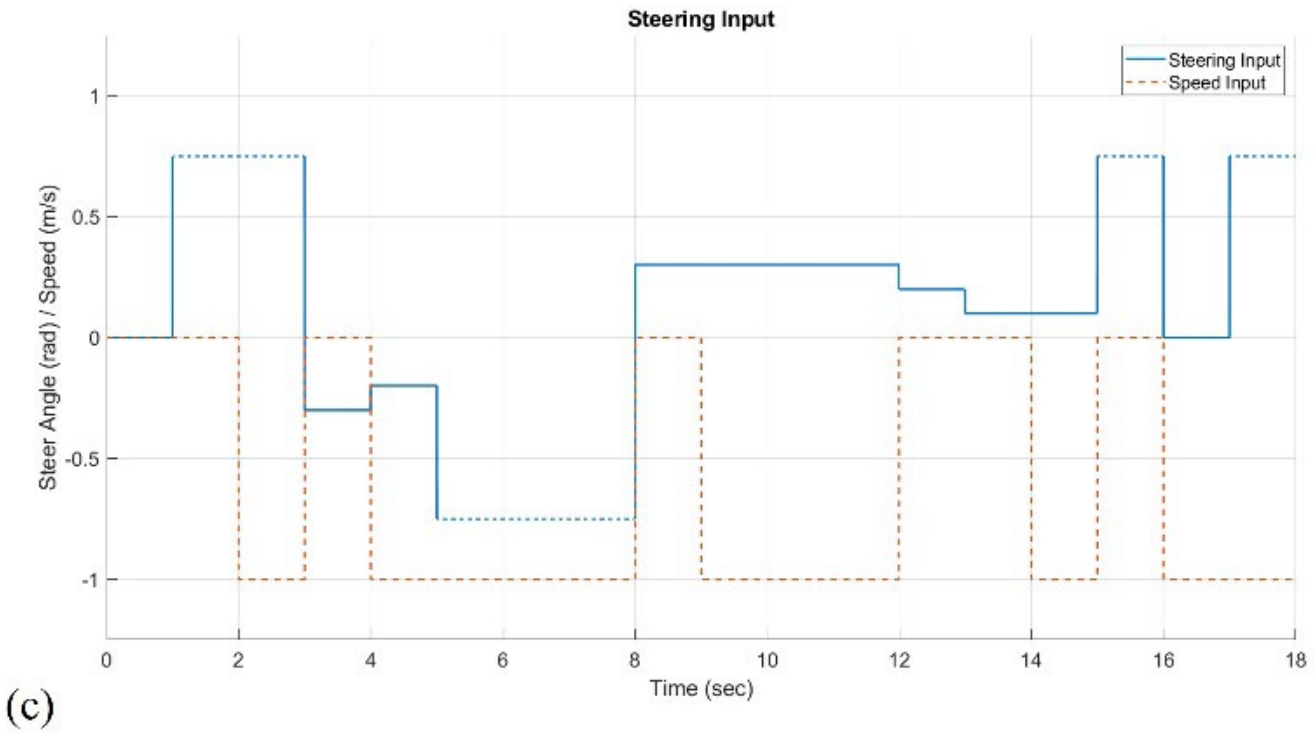

(c)

Figure 11. Advisory system results for $L_T = 2.5m$: (a) tractor-trailer trajectory; (b) hitch angle; (c) user inputs

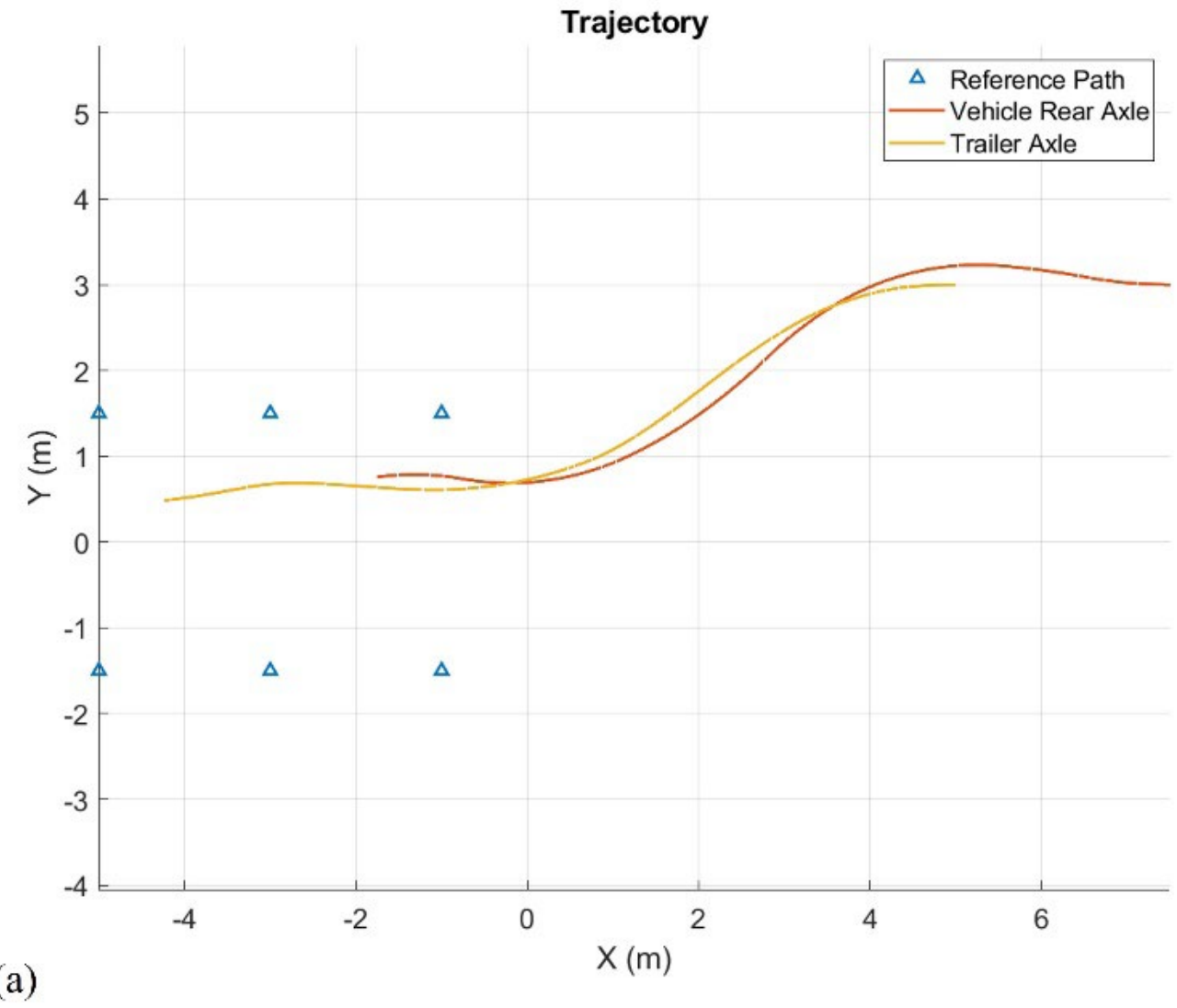

(a)

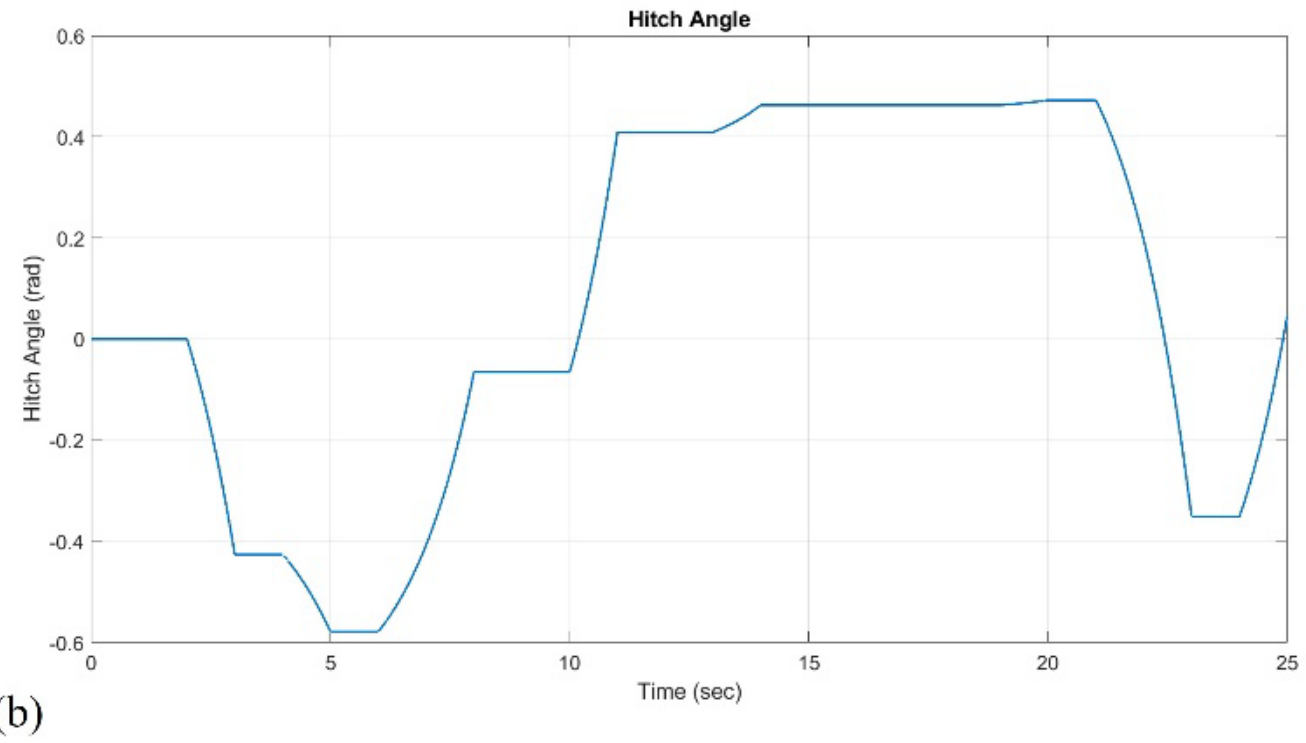

(b)

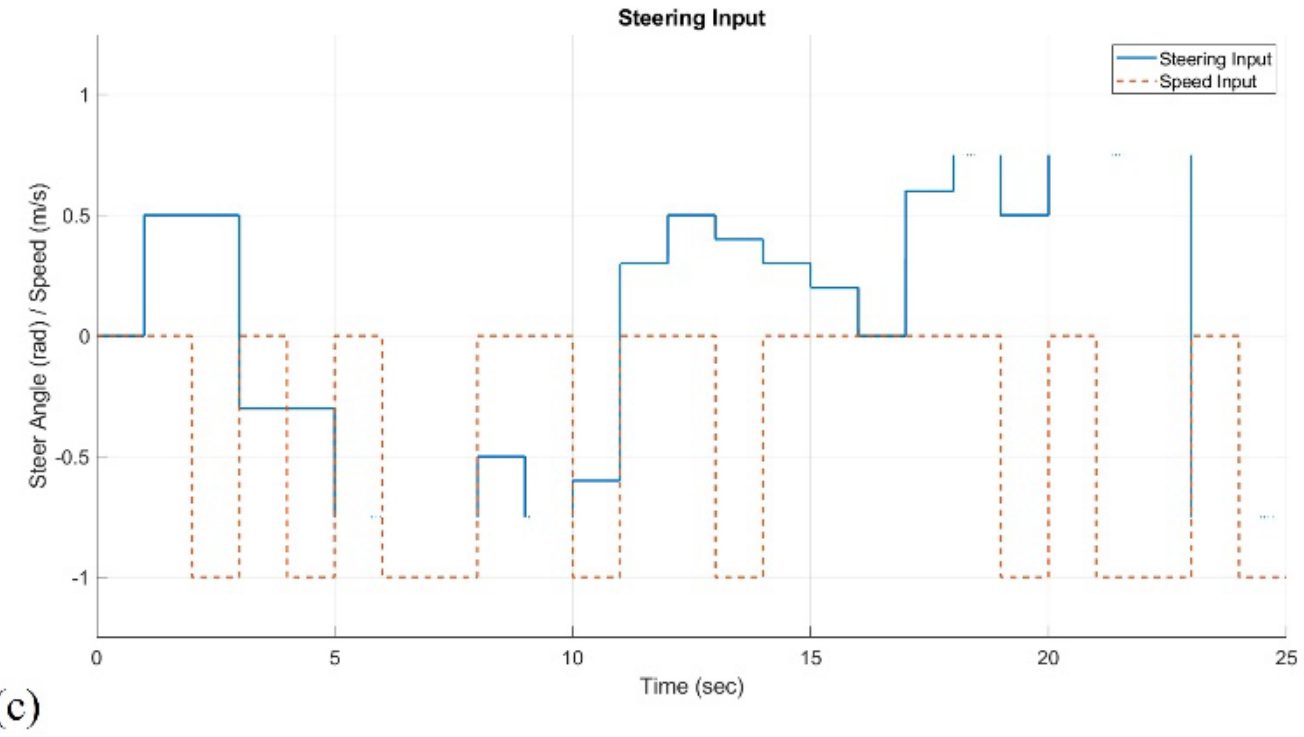

(c)

Figure 12. Advisory system results for $L_T = 1.5m$: (a) tractor-trailer trajectory; (b) hitch angle; (c) user inputs

## Conclusions

This paper presented an advisory system to help human drivers tackle vehicle-trailer system backing maneuvers. A kinematic vehicle-trailer model was derived to describe system behaviors at low speed, and inverse kinematics was also introduced to establish the mapping between a 'virtual' steer angle at the trailer hitch and the actual steer angle at the vehicle steerable axle. Simulation study showed that the location of the trailer hitch in relation to the vehicle rear axle affects how well the vehicle-trailer system can generate the 'virtual' steer angle required to orientate the trailer as desired. Advisory graphics was constructed using the inverse kinematics to display the current vehicle-trailer system configuration as well as predicted future behaviors of the trailer. Simulation and animation results confirm that this advisory system allows the human driver to carry out the backing maneuvers successfully with piecewise-continuous control inputs that resemble how these maneuvers are done in reality. The future work includes automating the vehicle-trailer backing maneuver as well as more simulations and experiments.

Future work can focus on the forward direction motion as well as the reverse motion treated here and cooperative driving [25], collision avoidance with vulnerable road users (VRU) and other vehicles [26-29] and V2X communication to help with perception of VRUs and other obstacles [30].

## Contact Information

Xincheng Cao: cao.1375@osu.edu

## Acknowledgments

The Ohio State University authors would like to thank HATCI (Hyundai America Technical Center, Inc.) for supporting this work.